\documentclass[runningheads]{llncs}
\usepackage{graphicx}
\usepackage{amsmath,amssymb} 
\usepackage{color}
\usepackage{multirow}
\usepackage{subcaption}
\usepackage{caption}
\usepackage[width=122mm,left=12mm,paperwidth=146mm,height=193mm,top=12mm,paperheight=217mm]{geometry}

\begin{document}

\thispagestyle{headings}
\pagestyle{plain}
\mainmatter
\def\ECCV18SubNumber{3DRMS - 004}  

\title{3D Depthwise Convolution: Reducing Model Parameters in 3D Vision Tasks } 



\author{Rongtian Ye$^{1} \qquad $Fangyu Liu$^{2,3}$ \qquad Liqiang Zhang$^{1}$\thanks{Corresponding author}}
\institute{
$^1$ Beijing Normal University   \quad
$^2$ University of Waterloo   \quad
$^3$ EPFL\\
\email{carmoac@mail.bnu.edu.cn\\
fangyu.liu@\{uwaterloo.ca,epfl.ch\}\\
zhanglq@bnu.edu.cn}
}


\newcommand{\chapternote}[1]{{%
  \let\thempfn\relax
  \footnotetext[0]{#1}
}}
\chapternote{Preprint. Work in progress.}

\maketitle

\begin{abstract}
Standard 3D convolution operations require much larger am\-ounts of memory and computation cost than 2D convolution operations. The fact has hindered the development of deep neural nets in many 3D vision tasks. In this paper, we investigate the possibility of applying depthwise separable convolutions in 3D scenario and introduce the use of 3D depthwise convolution. A 3D depthwise convolution splits a single standard 3D convolution into two separate steps, which would drastically reduce the number of parameters in 3D convolutions with more than one order of magnitude. We experiment with 3D depthwise convolution on popular CNN architectures and also compare it with a similar structure called pseudo-3D convolution. The results demonstrate that, with 3D depthwise convolutions, 3D vision tasks like classification and reconstruction can be carried out with more light-weighted neural networks while still delivering comparable performances.

\end{abstract}

\section{Introduction}

3D ConvNets have been widely used in almost all 3D computer vision tasks like 3D scene understanding (segmentation~\cite{tchapmi2017segcloud,liu20173dcnn}, classification~\cite{maturana2015voxnet,hegde2016fusionnet}, object detection~\cite{song2014sliding,liu20173dcnn}), human action / gesture recognition~\cite{ji20133d,molchanov2015hand}, video understanding~\cite{tran2015learning,qiu2017learning}, medical imaging~\cite{milletari2016v,cciccek20163d}, hyperspectral remote sensing images~\cite{li2017spectral}. It is powerful tool to extract high dimensional features from 3D data.

In real world scenarios, such as robotics, self-driving cars, augmented reality, computer-aided medical diagnosis, etc., the tasks need to be carried out on a platform with rather limited computing resources, constraining neural nets to small sizes. This led us to think about the possibility of using light-weighted 3D ConvNets with small amount of parameters. Besides, we have noticed that conventional researches have only explored relatively shallow 3D ConvNets, whose reason may relate to the great time and memory consumption of deep 3D ConvNets experiments. Computation cost is the bottleneck here. The problem becomes seeking a compromise (or balance) between 3D ConvNet's performance and its latency.

The number of parameters grows exponentially when convolution goes from 2D to 3D. For instance, a 2D convolution filter with shape $3\times 3$ only has $9$ parameters while a 3D one with the same side length, ie. a $3\times 3\times 3$ filter, would have $27$ parameters. With even greater side length of filters, the problem becomes much more severe: $125$ parameters there will be if the side length is $5$. Inspired by MobileNets~\cite{howard2017mobilenets}, which used the depthwise separable structure that factorized a standard 2D convolution into a depthwise convolution and a pointwise convolution (a $1\times 1$ convolution), we factorize a standard 3D convolution into: 1. separate convolutions on separate channels; 2. pointwise convolutions on all the channels. In most cases, the number of parameters in convolutional layers is reduced more than $10$ times. We do a formal analysis of this in math in section~\ref{2.1} to compare the number of parameters in standard 3D convolution and 3D depthwise convolution. 

In fact, there are already works in 3D vision community trying to reduce 3D convolution parameters with some success. To tackle the problem of using 3D ConvNets to learn video spatio-temporal representations (on a large scale), ~\cite{qiu2017learning} proposed an architecture to reduce reasonably many number of parameters by splitting one $3\times 3\times 3$ convolution into a $3\times 3\times 1$ convolution and a $1\times 1\times 3$ convolution. It is referred as the pseudo-3D convolution. Our work shows a different approach to decompose a standard 3D convolution operation. We compare these two architectures in terms of number of parameters (in section~\ref{2.2}) and performance (in experiments). The results indicate that 3D depthwise convolution reaches comparable performance with even fewer parameters.

We also experiment 3D depthwise convolution on some off-the-shelf CNN architectures like VGG~\cite{DBLP:journals/corr/SimonyanZ14a} and residual block~\cite{he2016deep}. In classification, it is shown that the number of parameters can be significantly reduced with little influence on neural nets' ability of extracting features. Besides classification, 3D depthwise convolution is open to be used in other tasks. We show an example by using 3D depthwise convolution in 3D reconstruction. Using 3D depthwise convolution in reconstruction networks brings several benefits: the original 3D vision tasks now can be carried out with fewer parameters (ie. models of smaller size); more complex and deeper models with too many parameters are now applicable due to reduction of parameters by 3D depthwise convolution. In experiments, a deeper 3D depthwise ConvNet decoder with fewer parameters beat a relatively shallow standard 3D ConvNet decoder with more parameters.


To conclude, the major contributions of this paper would be:
\begin{itemize}
\item proposes the use of 3D depthwise convolution to reduce parameters in 3D CNN models and analyzes the computing cost of 3D depthwise convolution in math, along with standard 3D convolution and pseudo-3D convolution;
\item investigates the performance of 3D depthwise convolution on popular CNN models and does extensive experiments adopting 3D depthwise convolution in 3D reconstruction networks.
\end{itemize}

\section{Proposed method}

In section~\ref{2.1} we formally introduce the 3D depthwise operation. In section~\ref{2.2}, we state the difference of pseudo-3D and 3D depthwise, then analyze why 3D depthwise would have less parameters. In section~\ref{2.3} we explain the use of 3D depthwise in reconstruction task and present the details of reconstruction networks we use.


\subsection{3D depthwise convolution}\label{2.1}
\begin{figure}[h]
\centering
\includegraphics[height=6.5cm]{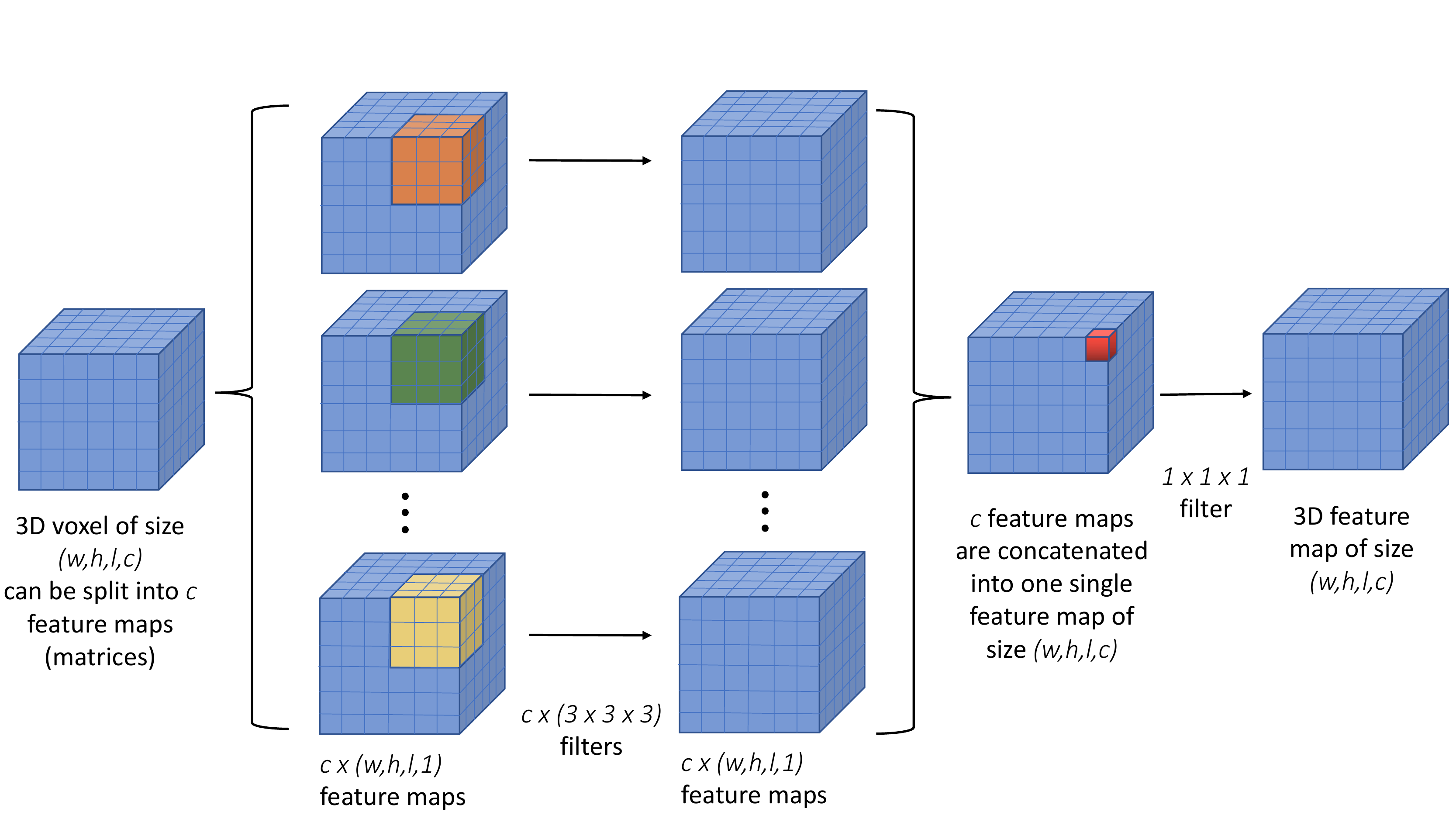}
\caption{We split the original 3D voxel grids (or 3D feature matrix) by the last dimension and get $c$ feature maps. Notice that $c$ is the number of channels. Then we use separate filters on separate feature maps. In the figure, it's $c$ separate $3\times 3\times 3$ filters on $c$ feature maps. After the filters, we stack all the results (ie. $c$ feature maps) by last dimension and get a feature map of the same size as the input (assuming paddings). And then we do a one-to-one convolution with filters of size $1\times 1 \times 1$. It is essentially a 3D fully-connected layer that does an element-wise linear combination on channels at very single voxel. This step is analogous to the \emph{pointwise} convolution in MobileNets.}
\label{fig:3ddw}
\end{figure}

\subsubsection{Standard 3D convolution} 
Given a 3D feature matrix with shape $(l,w,h,c)$, where $l,w,h$ represents length, width, height and $c$ denotes channels, the natural way of doing convolution operation on it would be using a filter with size $ k\times k\times k$ where $k$ is the side length of filter, to go over the 3D matrix. 

More formally, a standard convolution layer takes an input feature matrix $F$ with shape $(l_F,w_F,h_F,c_F)$ and outputs a feature matrix $G$ of size $(l_G,w_G,h_G,c_G)$. Notice that $c_F,c_G$ are number of channels before and after the convolution. The kernel $K$ here should be of size $k\times k \times k \times c_F \times c_G$ where $k$ is side length of the filter.

The output feature matrix for a standard 3D convolution is computed as 
$$G_{x,y,z,n} = \sum\limits_{i,j,k,m} K_{i,j,k,m,n}F_{x+i-1,y+j-1,z+k-1,m}$$
where $x,y,z$ and $i,j,k$ denotes voxel's spatial position and $m$ denotes the input channels while $n$ denotes the output channels.

And the computation cost would be: $$k \cdot k \cdot k \cdot c_F \cdot c_G \cdot l_F \cdot w_F \cdot h_F \ (1)$$

\subsubsection{3D depthwise convolution} 
In 3D depthwise convolution, we decompose one 3D convolution operation into two steps, using two filters: first, apply separate filters for each individual channel ; second, use $1\times 1\times 1\times c$ filter to apply an pointwise linear combination on feature maps output by the first step. Notice that we apply batchnorm~\cite{ioffe2015batch} and ReLU~\cite{le2015simple} after both steps.

The output feature matrix for a 3D depthwise convolution is computed as 
$$\hat{G}_{x,y,z,m} = \sum\limits_{i,j,k,m} \hat{K}_{i,j,k,m}F_{x+i-1,y+j-1,z+k-1,m}$$
where $x,y,z$ and $i,j,k$ again denotes the spatial position of a voxel. $\hat{K}$ is a depthwise convolution kernel of size $k\times k\times k\times c$ (consisting of $c$ filters). The $m$-th filter in $\hat{K}$ would be applied to the $m$-th channel in $F$. And the output of $m$-th filter becomes the $m$-th layer in $\hat{G}$. It is thus called depthwise / channelwise.

The computation cost would be: $$k \cdot k \cdot k \cdot c_F \cdot l_F \cdot w_F \cdot h_F $$

After each channel is depthwisely filtered, it remains to combine them into a single new feature map. We adopt one more pointwise convolution, ie. a $1\times 1\times 1$ convolution at every position of feature maps, to carry out a linear combination of layers of all depths. It is essentially fusing the splitted channels back together and activating exchange of information adequately across channels.

The pointwise convolution has a cost of:
$$c_F \cdot c_G \cdot l_F \cdot w_F \cdot h_F$$

And combining two steps together we get a cost of:
$$k \cdot k \cdot k \cdot c_F \cdot l_F \cdot w_F \cdot h_F + c_F \cdot c_G \cdot l_F \cdot w_F \cdot h_F \ (2)$$

\subsubsection{Comparing standard 3D convolution and 3D depthwise convolution} 
We get a reduction by using (2) dividing (1):
$$\frac{k \cdot k \cdot k \cdot c_F \cdot l_F \cdot w_F \cdot h_F + c_F \cdot c_G \cdot l_F \cdot w_F \cdot h_F}{k \cdot k \cdot k \cdot c_F \cdot c_G \cdot l_F \cdot w_F \cdot h_F}$$
$$ = \frac{k \cdot k \cdot k \cdot c_F +c_F\cdot c_G}{k \cdot k \cdot k \cdot c_F \cdot c_G} = \frac{1}{c_G}+\frac{1}{k^3} \ (\ast)$$

Channel size $c_G$ is empirically speaking a large number (usually $32, 64, 128,$ etc.) which makes $\frac{1}{c_G}$ very small. $\frac{1}{k^3}$ is depending on the side length of kernel. Even when the kernel is small and of side length $2$, $\frac{1}{k^3}$ is approaching to $0.1$ already. Combining the two very small terms, $(\ast)$ is easy to get to less than $0.1$ which means achieving a more than 10 times parameters reduction.

\subsection{Difference from pseudo-3D convolution}\label{2.2}
Using similar paradigm, we can analyze pseudo-3D convolution.

\begin{figure}[h]
\centering
\includegraphics[height=3cm]{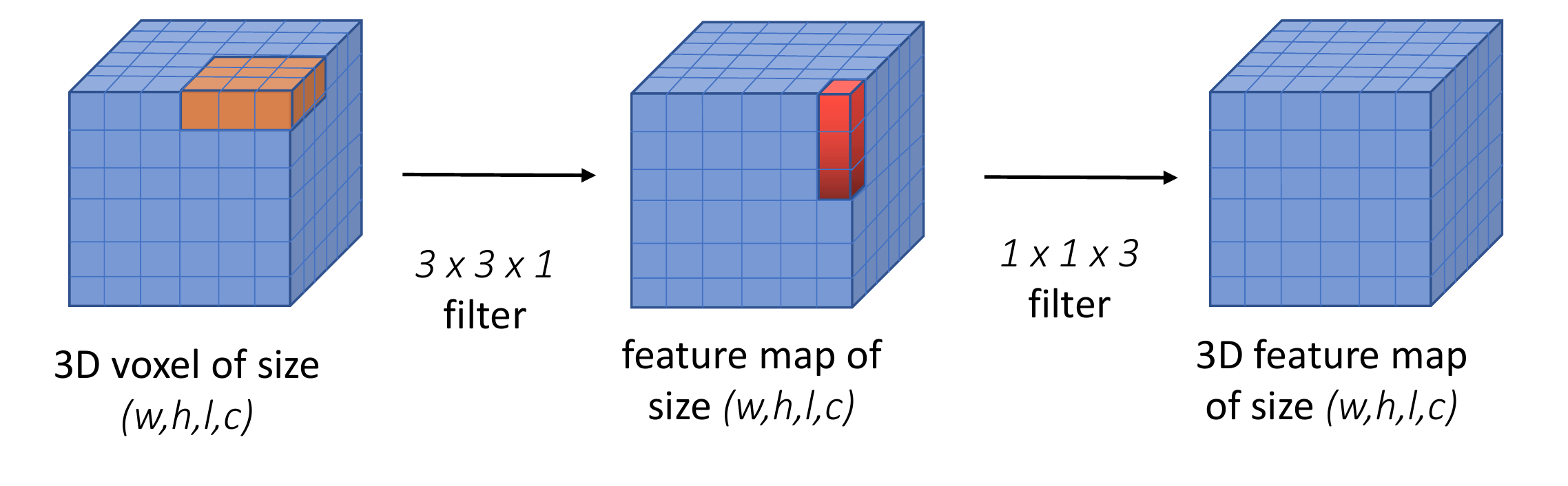}
\caption{The orange filter of size  $1\times 3 \times 3$ does a horizontal convolution (step one) and the red filter of size $3\times 1\times 1$ does a vertical convolution (step 2). Combining them together we have the pseudo-3D convolution.}
\label{fig:pseudo}
\end{figure}


Both pseudo-3D and 3D depthwise convolutions are splitting one single standard 3D convolution into two separate convolutions. But the splitting philosophy is different as suggested in figure~\ref{fig:3ddw} and figure~\ref{fig:pseudo}, which leads to very different number of parameters and behavior.



The first step of pseudo-3D convolution, which is basically a horizontal convolution, can be computed as 
$$\hat{G}_{x,y,z,m} = \sum\limits_{i,j,m} \hat{K}_{i,j,m,m}F_{x+i-1,y+j-1,z,m}$$

where $x,y,z,i,j,m$ are all analogous to previous definitions.

It has a computation cost of:
$$k\cdot k\cdot c_F \cdot c_F \cdot l_F\cdot w_F\cdot h_F$$

The second step can be seen as a vertical convolution can be computed as:
$$G_{x,y,z,n} = \sum\limits_{k,m} \hat{K}_{k,m,n}F_{x,y,z+k-1,m}$$

It has a computation cost of:
$$k\cdot c_F \cdot c_G \cdot l_F\cdot w_F\cdot h_F$$

The total computation cost of a pseudo-3D convolution is:
$$k\cdot k\cdot c_F \cdot c_F \cdot l_F\cdot w_F\cdot h_F + k\cdot c_F \cdot c_G \cdot l_F\cdot w_F\cdot h_F\ (3)$$

To compare the parameters of 3D depthwise convolution and pseudo-3D convolution, we divide (2) by (3):
$$\frac{k \cdot k \cdot k \cdot c_F \cdot l_F \cdot w_F \cdot h_F + c_F \cdot c_G \cdot l_F \cdot w_F \cdot h_F}{k\cdot k\cdot c_F \cdot c_F \cdot l_F\cdot w_F\cdot h_F + k\cdot c_F \cdot c_G \cdot l_F\cdot w_F\cdot h_F}$$
$$= \frac{k^3 + c_G}{k^2\cdot c_F + k\cdot c_G} \approx \frac{k}{c_F}\ (\star)$$

Comparing to channel size $c_F$, kernel side length $k$ is usually really small (eg. mostly likely to be $3$ or $5$ or $7$). 3D depthwise, who is breaking the chain of multiplication by depth / channel, gains an edge over pseudo-3D, who is breaking the chain by kernel side length.

We also compare our proposed method with the pseudo 3D convolution idea in experiments, finding that the 3D depthwise achieves comparable performance with even fewer parameters.

\subsection{Use 3D depthwise convolution in reconstruction networks}\label{2.3}

In this section, we state the use of 3D depthwise convolution in 3D reconstruction task.

\begin{figure}[h]
\centering
\includegraphics[height=5cm]{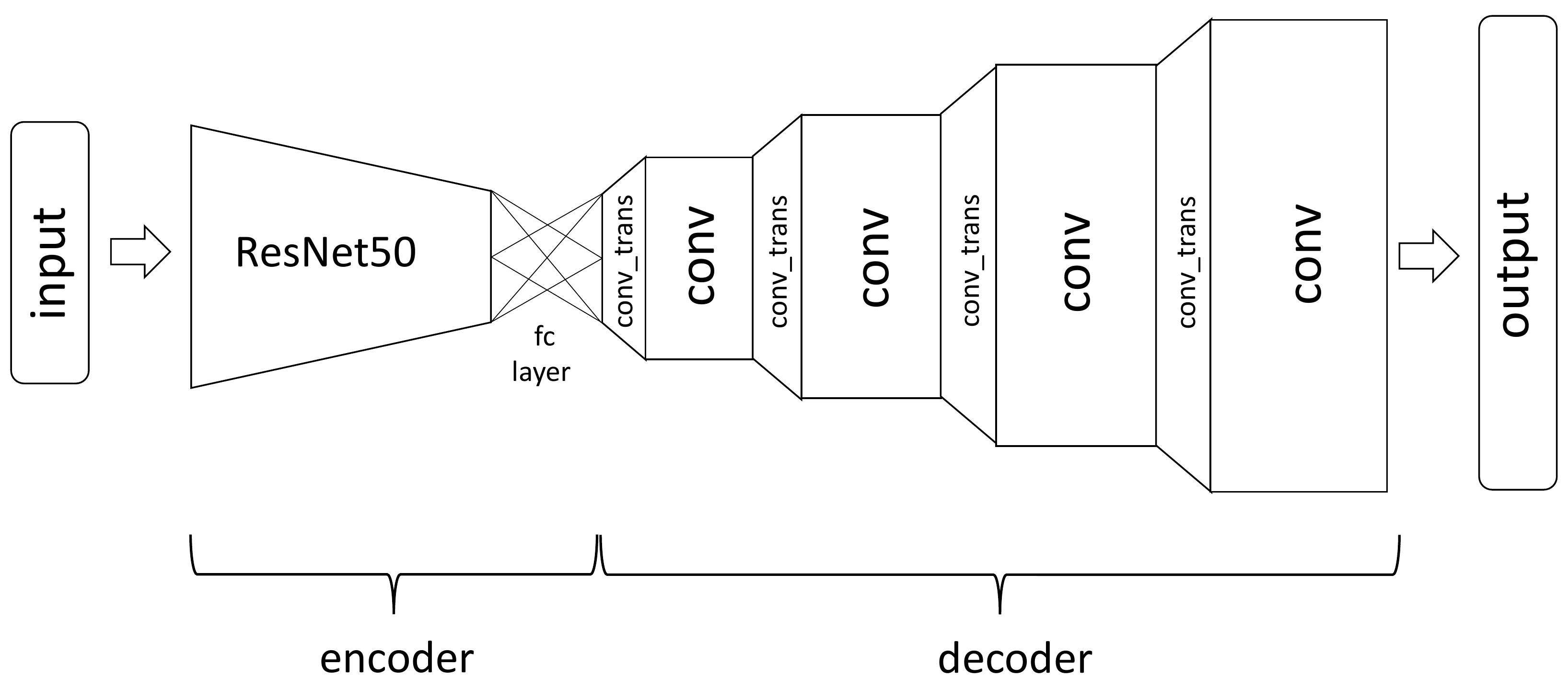}
\caption{A pipeline of the reconstruction networks.}
\label{fig:rec-pipe}
\end{figure}

3D reconstruction nets can be seen as autoencoders. For single view reconstruction, the input of neural net is an image, the output is a 3D voxel grids depicting the object. With (image,voxel grids) pairs as training data, we can do supervised learning on the autoencoder.

The autoencoder consists of an encoder, which is a 2D ConvNet to encode image into its vector representation, and an decoder, which is a upsampling 3D ConvNet to decode the encoded vector into 3D voxel grids. The general pipeline can be found in figure~\ref{fig:rec-pipe}.

In our setting, we use ResNet50 as the image encoder. Specifically, we extract the output of average pooling layer in ResNet50, which would be a 1-D vector with length 2048. Then, we use a fully-connected layer to map the vector into a 1024-length 1-D vector. We regard this vector as the output of encoder and input of decoder. Decoder is where we test the use of 3D depthwise convolution. The decoding steps would be: 
\begin{itemize}
\item use a transposed convolution kernel of size $4\times 4\times 4$ to map the 1-D vector to a feature matrix of size $(4,4,4,256)$ where $256$ is the number of channels;
\item do several similar transposed 3D convolution to map the feature matrix to the size of $8\times 8 \times 8 \times 128$, $16\times 16 \times 16 \times 64$, then $32\times 32 \times 32 \times 32$;
\item use kernel of size $1\times 1\times 1$ to reduce number of channels back to 1 and get the $32\times 32 \times 32 \times 1$ feature matrix we desired.
\end{itemize}

\begin{figure}[h]
\centering
\begin{subfigure}{.5\textwidth}
  \centering
  \includegraphics[width=.4\linewidth]{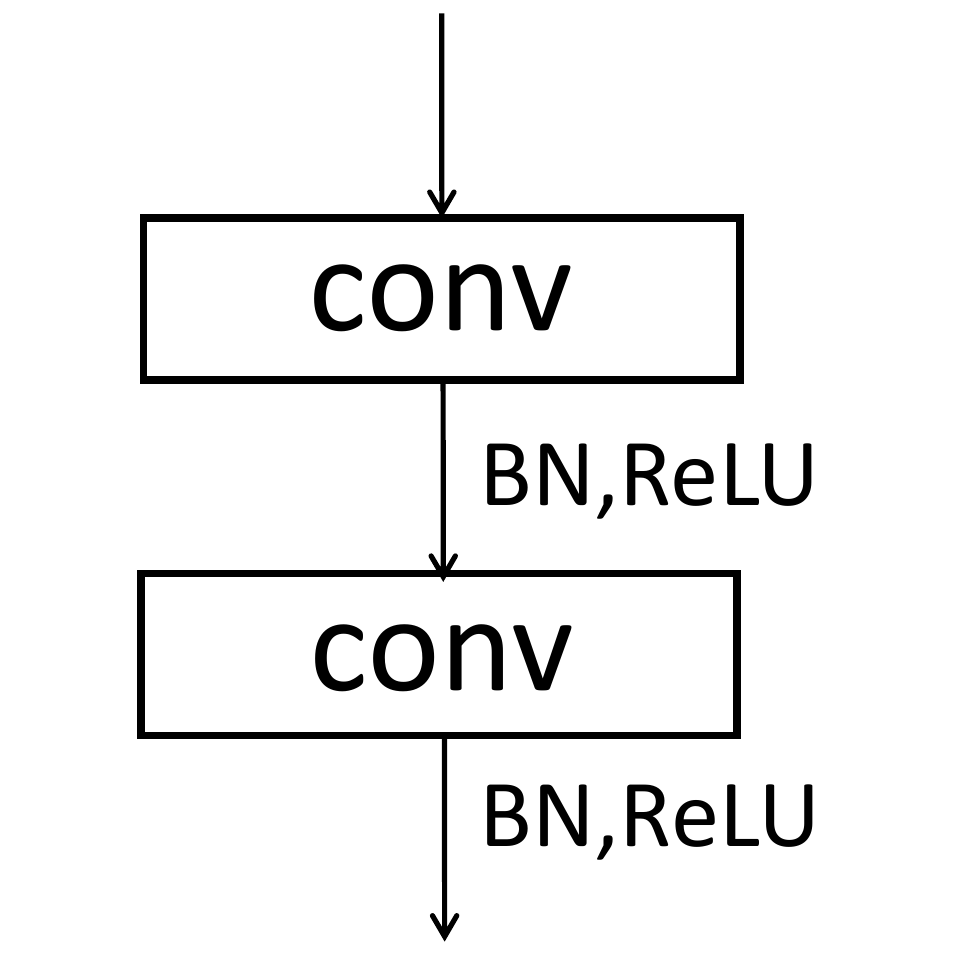}
  \caption{(a) Regular convolution block.}
  \label{fig:sub1}
\end{subfigure}%
\begin{subfigure}{.5\textwidth}
  \centering
  \includegraphics[width=.35\linewidth]{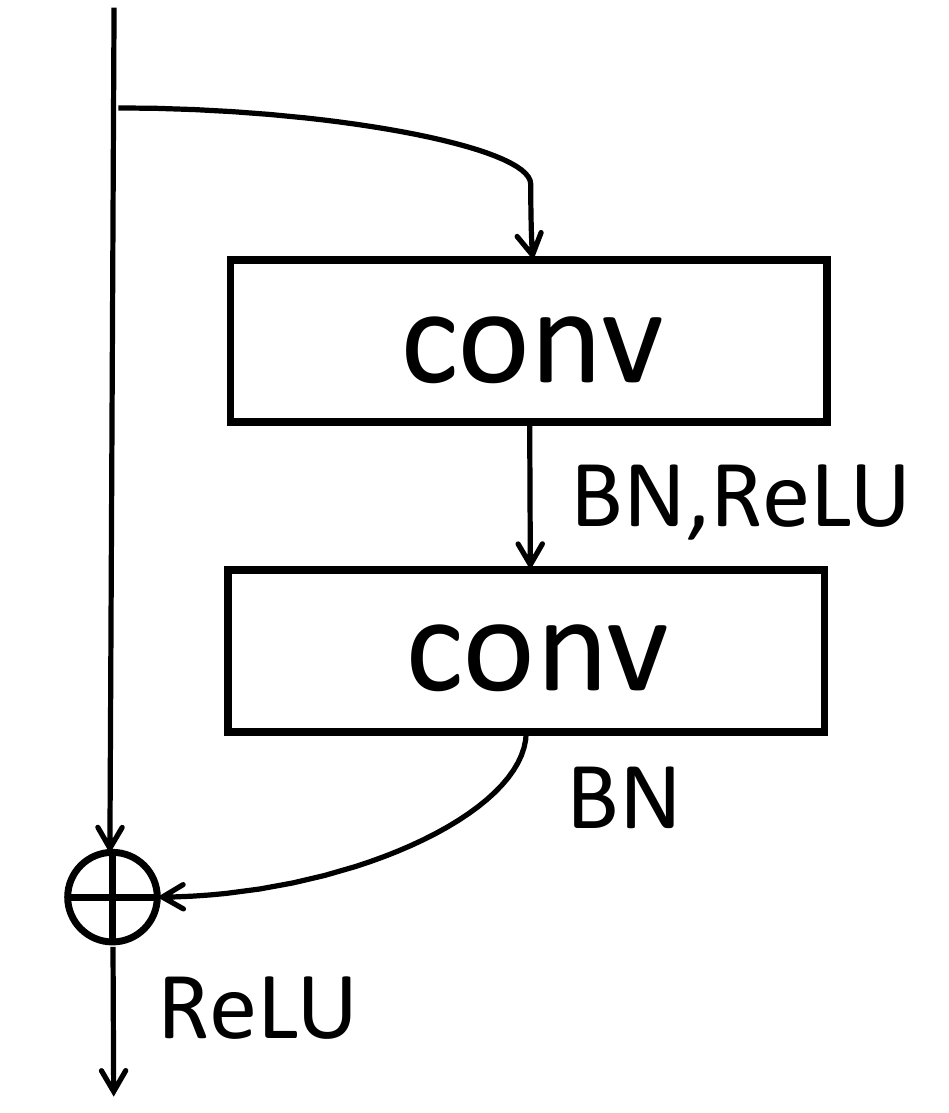}
  \caption{(b) Residual convolution block.}
  \label{fig:sub2}
\end{subfigure}
\caption{Two different blocks used in reconstruction networks. (a) is used in all Rec nets while (b) is used in all ResRec nets.}
\label{fig:block}
\end{figure}

For the decoder, we designed convolution blocks with and without residual structure as suggested in figure~\ref{fig:block}. We tested their performance by using standard 3D convolution, 3D depthwise convolution, as well as pseudo-3D convolution. Experiment results of comparing the combinations of \{residual block, regular block\}$\times$\{standard 3D convolution, pseudo-3D convolution, 3D depthwise convolution\} can be seen in section~\ref{3}.

\section{Experiments}\label{3}

\subsection{Dataset setup}
We used ShapeNetCore~\cite{chang2015shapenet} as our experiment dataset. To be consistent with some of our peers' works which we are comparing results with, we use the exact same dataset as~\cite{kar2017learning}. The set contains $44,000$ 3D models and $13$ categories with a train/val/test split of $[0.7,0.1,0.2]$.
We use the $64\times64\times 64$ resolution voxel grids on classification experiments and $32\times32\times 32$ ones on reconstruction.

\subsection{Classification}

We start with the most common task - classification. We compare the performance of standard 3D convolution, pseudo-3D convolution and 3D depthwise convolution by applying them on VGGs of different number of layers. 

\subsubsection{Training}
We train every model for 20 epochs. The first 10 uses a learning rate of ${10}^{-5}$ and the second 10 uses ${10}^{-6}$. Batch size is set to $8$. Models take 3D voxel grids of size $(64,64,64)$ as input and output a vector of length $13$ as prediction. Every element in the $(1,13)$ vector represents the probability of the input belongs to a corresponding class of object. And we use CrossEntropy as loss function.

\subsubsection{Results}
We found that standard 3D convolution, pseudo-3D convolution, 3D depthwise convolution's performances are comparable in this task. As suggested in table~\ref{table:classification}, the parameters of convolution layers are reduced by more than $95\%$ in all three VGGs. And the total numbers of parameters are reduced by $40\% - 60\%$. In these VGGs, the fully-connected layer takes $33,579,021$ parameters and contributes to most of the parameters in the whole network. Depthwise convolutional layers are only using very minimum parameters comparing to it.


\begin{table}[h]
\centering
\caption{Comparison of 3D depthwise convolution and standard 3D convolution on VGG in applications of classification task.}
\label{table:classification}
\begin{tabular}{l|c|cc|ccll}
\cline{1-6}
\multirow{2}{*}{method} & \multirow{2}{*}{accuracy} & \multirow{2}{*}{\begin{tabular}[c]{@{}c@{}}\# param \\ in conv layers\end{tabular}} & \multirow{2}{*}{reduced by} & \multirow{2}{*}{\begin{tabular}[c]{@{}c@{}}\# parameters \\ in total\end{tabular}} & \multirow{2}{*}{reduced by} &  &  \\
                        &                           &                                                                                     &                             &                                                                                    &                             &  &  \\ \cline{1-6}
3D VGG13                & 95.11                   & 28,261,824                                                                          &                             & 61,840,845                                                                         &                             &  &  \\
pseud-3D VGG13          & 95.12                   & 11,823,819                                                                          & 58.16\%                     & 45,402,840                                                                         & 26.58\%                     &  &  \\
3D dw VGG13             & 95.10                   & 1,174,237                                                                           & \textbf{95.85\%}                     & 34,753,258                                                                         & \textbf{43.80\%}                     &  &  \\ \cline{1-6}
3D VGG16                & 95.21                   & 44,189,632                                                                          &                             & 77,768,653                                                                         &                             &  &  \\
pseud-3D VGG16          & 94.93                   & 18,906,827                                                                          & 57.21\%                     & 52,485,848                                                                         & 32.51\%                     &  &  \\
3D dw VGG16             & 94.26                   & 1,803,741                                                                           & \textbf{95.92\%}                     & 35,382,762                                                                         & \textbf{54.50\%}                     &  &  \\ \cline{1-6}
3D VGG19                & 94.95                   & 60,117,440                                                                          &                             & 93,696,461                                                                         &                             &  &  \\
pseud-3D VGG19          & 94.61                   & 25,989,835                                                                          & 56.77\%                     & 59,568,856                                                                         & 36.42\%                     &  &  \\
3D dw VGG19             & 94.71                   & 2,433,245                                                                           & \textbf{95.95\%}                     & 36,012,266                                                                         & \textbf{61.56\%}                     &  &  \\ \cline{1-6}
\end{tabular}
\end{table}

\subsection{3D Reconstruction}

Previous works that focus on 3D reconstruction usually use several regular convolutions together with transpose convolutions in decoders to map encoded vectors back to 3D voxel~\cite{girdhar2016learning,wu2016learning,yan2016perspective}. The major constraint of applying more complex structure here is that 3D ConvNets have plenty of parameters, which makes deepening them very hard. For instance, to enhance the capability of decoder (the mapping from image encodings to 3D voxel), 3D-R2N2~\cite{choy20163d} used a residual block based architecture. However, due to the massive amount of parameters in the 3D neural nets, its depth (only 10 layers) is still not enough comparing to ResNets in 2D scenarios~\cite{he2016deep}. So, reducing number of parameters is without doubt the key to come up with a better decoder here.

Fig~\ref{fig:rec-pipe} shows the pipeline of the reconstruction network we are experimenting on. We use identical encoders (ResNet50) for all experiments and test different decoder architectures, both with and without 3D depthwise, as suggested in table~\ref{table:reconstruction_arch}. 

\subsubsection{Metrics}
We use the mean Intersection-over-Union (mIoU) between ground truth and the models' predicted 3D voxel grids to evaluate performance of 3D reconstruction. Specifically, for an individual voxel grid, the IoU is defined as

$$mIoU = \frac{\sum_{i,j,k}[I(p_{(i,j,k)>t})\cdot I(y_{(i,j,k)})]}{\sum_{i,j,k}[I(I(p_{(i,j,k)>t})+I(y_{(i,j,k)}))]} \ \ (\ast\ast)$$
where $I(\cdot)$ is an indicator function, $y_{(i,j,k)}$ is the groundtruth at position $(i,j,k)$ while $p_{(i,j,k)}$ is the prediction of it and $t$ is a threshold. The higher the mIoU, the better the 3D reconstruction model is.

\subsubsection{Training}
We train every model for 120 epochs. The first 60 uses a learning rate of ${10}^{-6}$ and the second 60 uses ${10}^{-7}$. Batch size is set to 32. Models take images of size $(224,224)$ as input and output prediction (3D voxel) of size $(32,32,32)$. We use voxel-wise CrossEntropy as loss function. To evaluate a model, we set a threshold for binarizing every voxel, ie. choosing $t$ in equation $(\ast\ast)$. We tested every one of $\{0.1,0.3,0.5,0.7,0.9\}$ and found that models generally have the highest mIoU when $t=0.3$. For consistency, all experiments are using $t=0.3$ as threshold.
\begin{table}[h]
\caption{Architectures of reconstruction nets. Parameters written in the format of (kernel\_size,output\_channel\_size,stride). $\ast$ means it's residual block.}
\label{table:reconstruction_arch}

\begin{tabular}{c|c|cccc}
\hline
layer                  & output                         & \multicolumn{1}{c|}{Rec-6}                                                                                                                      & \multicolumn{1}{c|}{ResRec-6}                                                                                                                         & \multicolumn{1}{c|}{Rec-16}                                                                                                                   & ResRec-16                                                                                                                        \\ \hline
encoder                & (1,1,1,2048)                   & \multicolumn{4}{c}{ResNet50 (avgpool layer out)}                                                                                                                                                                                                                                                                                                                                                                                                                                                                                                                                           \\ \hline
fc                     & (1,1,1,1024)                   & \multicolumn{4}{c}{fully connected layer}                                                                                                                                                                                                                                                                                                                                                                                                                                                                                                                                                  \\ \hline
convtrans1             & \multirow{4}{*}{(4,4,4,256)}   & \multicolumn{4}{c}{conv\_transpose $4\times 4\times 4, 256$, stride 1}                                                                                                                                                                                                                                                                                                                                                                                                                                                                                                                     \\ \cline{1-1} \cline{3-6} 
\multirow{3}{*}{conv1} &                                & \multicolumn{1}{c|}{\multirow{3}{*}{\begin{tabular}[c]{@{}c@{}}$[3\times3\times3,256;$\\ $3\times3\times3,256]$\\ $\times 1$\end{tabular}}}     & \multicolumn{1}{c|}{\multirow{3}{*}{\begin{tabular}[c]{@{}c@{}}$[3\times3\times3,256;$\\ $3\times3\times3,256]$\\ $\times 1$, $\ast$\end{tabular}}}   & \multicolumn{1}{c|}{\multirow{3}{*}{\begin{tabular}[c]{@{}c@{}}$[3\times3\times3,256;$\\ $3\times3\times3,256]$\\ $\times 2$\end{tabular}}}   & \multirow{3}{*}{\begin{tabular}[c]{@{}c@{}}$[3\times3\times3,256;$\\ $3\times3\times3,256]$\\ $\times 2$, $\ast$\end{tabular}}   \\
                       &                                & \multicolumn{1}{c|}{}                                                                                                                           & \multicolumn{1}{c|}{}                                                                                                                                 & \multicolumn{1}{c|}{}                                                                                                                         &                                                                                                                                  \\
                       &                                & \multicolumn{1}{c|}{}                                                                                                                           & \multicolumn{1}{c|}{}                                                                                                                                 & \multicolumn{1}{c|}{}                                                                                                                         &                                                                                                                                  \\ \hline
convtrans2             & \multirow{4}{*}{(8,8,8,128)}   & \multicolumn{4}{c}{conv\_transpose $2\times 2\times 2$, 128, stride 2}                                                                                                                                                                                                                                                                                                                                                                                                                                                                                                                     \\ \cline{1-1} \cline{3-6} 
\multirow{3}{*}{conv2} &                                & \multicolumn{1}{c|}{\multirow{3}{*}{\begin{tabular}[c]{@{}c@{}}$[3\times 3\times 3,128;$\\ $3\times 3\times 3,128]$\\ $\times 1$\end{tabular}}} & \multicolumn{1}{c|}{\multirow{3}{*}{\begin{tabular}[c]{@{}c@{}}$[3\times3\times3,128;$\\ $3\times3\times3,128]$\\ $\times 1$, $\ast$\end{tabular}}}   & \multicolumn{1}{c|}{\multirow{3}{*}{\begin{tabular}[c]{@{}c@{}}$[3\times3\times3,128;$\\ $3\times3\times3,128]$\\ $\times 2$\end{tabular}}}   & \multirow{3}{*}{\begin{tabular}[c]{@{}c@{}}$[3\times3\times3,128;$\\ $3\times3\times3,128]$\\ $\times 2$, $\ast$\end{tabular}}   \\
                       &                                & \multicolumn{1}{c|}{}                                                                                                                           & \multicolumn{1}{c|}{}                                                                                                                                 & \multicolumn{1}{c|}{}                                                                                                                         &                                                                                                                                  \\
                       &                                & \multicolumn{1}{c|}{}                                                                                                                           & \multicolumn{1}{c|}{}                                                                                                                                 & \multicolumn{1}{c|}{}                                                                                                                         &                                                                                                                                  \\ \hline
convtrans3             & \multirow{4}{*}{(16,16,16,64)} & \multicolumn{4}{c}{conv\_transpose $2\times 2\times 2$, 64, stride 2}                                                                                                                                                                                                                                                                                                                                                                                                                                                                                                                      \\ \cline{1-1} \cline{3-6} 
\multirow{3}{*}{conv3} &                                & \multicolumn{1}{c|}{\multirow{3}{*}{\begin{tabular}[c]{@{}c@{}}$[3\times 3\times 3,64;$\\ $3\times 3\times 3,64]$\\ $\times 1$\end{tabular}}}   & \multicolumn{1}{c|}{\multirow{3}{*}{\begin{tabular}[c]{@{}c@{}}$[3\times 3\times 3,64;$\\ $3\times 3\times 3,64]$\\ $\times 1$, $\ast$\end{tabular}}} & \multicolumn{1}{c|}{\multirow{3}{*}{\begin{tabular}[c]{@{}c@{}}$[3\times 3\times 3,64;$\\ $3\times 3\times 3,64]$\\ $\times 2$\end{tabular}}} & \multirow{3}{*}{\begin{tabular}[c]{@{}c@{}}$[3\times 3\times 3,64;$\\ $3\times 3\times 3,64]$\\ $\times 2$, $\ast$\end{tabular}} \\
                       &                                & \multicolumn{1}{c|}{}                                                                                                                           & \multicolumn{1}{c|}{}                                                                                                                                 & \multicolumn{1}{c|}{}                                                                                                                         &                                                                                                                                  \\
                       &                                & \multicolumn{1}{c|}{}                                                                                                                           & \multicolumn{1}{c|}{}                                                                                                                                 & \multicolumn{1}{c|}{}                                                                                                                         &                                                                                                                                  \\ \hline
convtrans4             & \multirow{4}{*}{(32,32,32,32)} & \multicolumn{4}{c}{conv\_transpose $2\times 2\times 2$, 32, stride 2}                                                                                                                                                                                                                                                                                                                                                                                                                                                                                                                      \\ \cline{1-1} \cline{3-6} 
\multirow{3}{*}{conv4} &                                & \multicolumn{1}{c|}{\multirow{3}{*}{}}                                                                                                          & \multicolumn{1}{c|}{\multirow{3}{*}{}}                                                                                                                & \multicolumn{1}{c|}{\multirow{3}{*}{\begin{tabular}[c]{@{}c@{}}$[3\times 3\times 3,32;$\\ $3\times 3\times 3,32]$\\ $\times 2$\end{tabular}}} & \multirow{3}{*}{\begin{tabular}[c]{@{}c@{}}$[3\times 3\times 3,32;$\\ $3\times 3\times 3,32]$\\ $\times 2$, $\ast$\end{tabular}} \\
                       &                                & \multicolumn{1}{c|}{}                                                                                                                           & \multicolumn{1}{c|}{}                                                                                                                                 & \multicolumn{1}{c|}{}                                                                                                                         &                                                                                                                                  \\
                       &                                & \multicolumn{1}{c|}{}                                                                                                                           & \multicolumn{1}{c|}{}                                                                                                                                 & \multicolumn{1}{c|}{}                                                                                                                         &                                                                                                                                  \\ \hline
conv5                  & (32,32,32,1)                   & \multicolumn{4}{c}{$1\times 1\times 1$, 1, stride 1}                                                                                                                                                                                                                                                                                                                                                                                                                                                                                                                                       \\ \hline
\end{tabular}

\end{table}

\setlength{\tabcolsep}{4pt}
\begin{table}[h]
\begin{center}
\caption{comparison of 3d depthwise conv and regular 3d conv in reconstruction networks}
\label{table:reconstruction}
\end{center}
\end{table}
\setlength{\tabcolsep}{1.4pt}

\subsubsection{Results}

Table~\ref{table:miou-param} shows that: 
\begin{itemize}
\item With same number of layers, 3D depthwise looses $\sim 2.2\%$ mIoU on average while pseudo-3D looses $\sim 1.7\%$ on average. But 3D depthwise reduces significantly more parameters than pseudo-3D.
\item A deeper 3D FCN with 3D depthwise convolution (ResRec-16 dw) achieves better accuracy with fewer parameters (\# params in decoder: $17,533,792$) than a shallow standard FCN (Rec-6) with more parameters (\# param in decoder: $21,768,928$) as suggested in table~\ref{table:miou-param}.
\end{itemize}
\begin{table}[h]
\centering
\begin{tabular}{l|c|cc|cc}
\hline
\multirow{2}{*}{method} & \multirow{2}{*}{mIoU} & \multirow{2}{*}{\begin{tabular}[c]{@{}c@{}}\# param in \\ conv layers\end{tabular}} & \multirow{2}{*}{reduced by} & \multirow{2}{*}{\begin{tabular}[c]{@{}c@{}}\# param in \\ decoder\end{tabular}} & \multirow{2}{*}{reduced by} \\
                        &                       &                                                                                     &                             &                                                                               &                             \\ \hline
Rec-6                   & 61.4                & 4,646,656                                                                           &                             & 21,768,928                                                                    &                             \\
Rec-6 pseudo            & 59.1                & 2,067,968                                                                           & 55.50\%                     & 19,190,240                                                                    & 11.85\%                     \\
Rec-6 dw                & 58.1                & 201,600                                                                             & \textbf{95.66\%}                     & 17,323,872                                                                    & \textbf{20.42}\%                     \\ \hline
ResRec-6                & 61.7                & 4,646,656                                                                           &                             & 21,768,928                                                                    &                             \\
ResRec-6 pseudo         & 61.0                & 2,067,968                                                                           & 55.50\%                     & 19,190,240                                                                    & 11.85\%                     \\
ResRec-6 dw             & 60.4                & 201,600                                                                             & \textbf{95.66\%}                     & 17,323,872                                                                    & \textbf{20.42\%}                     \\ \hline
Rec-16                  & 63.4                & 9,404,160                                                                           &                             & 26,526,432                                                                    &                             \\
Rec-16 pseudo           & 61.2                & 4,185,600                                                                           & 55.49\%                     & 21,307,872                                                                    & 19.67\%                     \\
Rec-16 dw               & 60.9                & 411,520                                                                             & \textbf{95.62\%}                     & 17,533,792                                                                    & \textbf{33.90\%}                     \\ \hline
ResRec-16               & 63.1                & 9,404,160                                                                           &                             & 26,526,432                                                                    &                             \\
ResRec-16 pseudo        & 61.7                & 4,185,600                                                                           & 55.49\%                     & 21,307,872                                                                    & 19.67\%                     \\
ResRec-16 dw            & 61.5                & 411,520                                                                             & \textbf{95.62\%}                     & 17,533,792                                                                    & \textbf{33.90\% }                   \\ \hline
\end{tabular}
\caption{Comparing reconstruction results of using different 3D convolutions in various decoder structures. ``Res'' means with residual block; ``dw'' is short for depthwise. }
\label{table:miou-param}
\end{table}

Table~\ref{table:rec-full-results} demonstrates the full quantitative results of different reconstruction networks, including two prior works and the combinations of \{residual block, regular block\}$\times$\{standard 3D convolution, pseudo-3D convolution, 3D depthwise convolution\}. We notice that Rec-16, who has the largest depth and most parameters, also has the best performance.

\begin{table}[h]
\centering
\scriptsize
\begin{tabular}{l|cccccccccccccc}
\hline
\multicolumn{1}{c|}{method} & mIoU                      & aero & bench & cabinet & car  & chair & display & lamp & speaker & rifle & sofa & table & phone & vessel \\ \hline
3D-R2N2~\cite{choy20163d}                     & \multicolumn{1}{c|}{55.1} & 56.7 & 43.2  & 61.8    & 77.6 & 50.9  & 44.0    & 40.0 & 56.7    & 56.5  & 58.9 & 51.6  & 65.6  & 53.1   \\
V-lsm~\cite{kar2017learning}                       & \multicolumn{1}{c|}{61.5} & 61.7 & 50.8  & 65.9    & 79.3 & 57.8  & 57.9    & 48.1 & 63.9    & 69.7  & 67.0 & 55.6  & 67.6  & 58.3   \\ \hline
Rec-6               & \multicolumn{1}{c|}{61.4} & 59.7 & 55.5  & 72.4    & 79.3 & 55.9  & 51.5    & 44.8 & 63.6    & 60.1  & 66.8 & 58.7  & 71.2  & 58.3   \\
Rec-6 pd            & \multicolumn{1}{c|}{59.1} & 56.9 & 52.6  & 69.5    & 76.4 & 54.2  & 49.7    & 44.5 & 63.4    & 58.1  & 65.6 & 56.2  & 65.4  & 55.4   \\
Rec-6 dw            & \multicolumn{1}{c|}{58.1} & 54.7 & 50.3  & 67.9    & 77.8 & 53.0  & 49.1    & 41.8 & 62.0    & 57.5  & 63.7 & 54.4  & 67.9  & 55.7   \\ \hline
ResRec-6            & \multicolumn{1}{c|}{61.7} & 60.7 & 57.3  & 71.9    & 79.2 & 56.8  & 52.5    & 44.9 & 62.1    & 60.2  & 68.2 & 59.1  & 70.9  & 58.4   \\
ResRec-6 pd         & \multicolumn{1}{c|}{61.0} & 59.2 & 54.5  & 71.8    & 79.5 & 55.5  & 50.6    & 44.9 & 63.8    & 59.8  & 67.2 & 57.9  & 69.9  & 57.9   \\
ResRec-6 dw         & \multicolumn{1}{c|}{60.4} & 57.8 & 52.5  & 71.1    & 78.9 & 55.1  & 51.5    & 43.5 & 63.3    & 59.7  & 66.1 & 56.9  & 71.7  & 57.7   \\ \hline
Rec-16              & \multicolumn{1}{c|}{\textbf{63.4}} & 63.3 & 59.2  & 74.2    & 80.0 & 57.7  & 54.4    & 46.7 & 64.7    & 62.0  & 69.5 & 60.0  & 74.1  & 58.1   \\
Rec-16 pd           & \multicolumn{1}{c|}{61.2} & 58.8 & 54.8  & 71.9    & 79.2 & 55.1  & 53.2    & 43.4 & 62.9    & 60.1  & 67.4 & 57.9  & 73.4  & 57.5   \\
Rec-16 dw           & \multicolumn{1}{c|}{60.9} & 60.5 & 52.9  & 71.2    & 79.6 & 55.5  & 50.6    & 40.6 & 65.8    & 61.3  & 68.4 & 57.2  & 70.5  & 58.3   \\ \hline
ResRec-16           & \multicolumn{1}{c|}{63.1} & 59.2 & 58.1  & 74.6    & 79.1 & 58.1  & 54.4    & 47.2 & 66.0    & 60.6  & 69.4 & 61.2  & 74.3  & 58.3   \\
ResRec-16 pd        & \multicolumn{1}{c|}{61.7} & 60.6 & 56.3  & 72.3    & 79.3 & 56.8  & 52.1    & 45.5 & 64.0    & 61.0  & 68.2 & 58.4  & 70.0  & 57.9   \\
ResRec-16 dw        & \multicolumn{1}{c|}{61.5} & 59.3 & 55.2  & 73.1    & 79.1 & 55.1  & 52.8    & 43.7 & 64.5    & 59.2  & 68.2 & 58.9  & 72.4  & 57.5   \\ \hline
\end{tabular}
\small
\caption{Full results of all kinds of reconstruction networks. ``Res'' means with residual block; ``dw'' is short for depthwise; ``pd'' is short for pseudo. }
\label{table:rec-full-results}
\end{table}

\begin{figure}[h!]
\centering
\includegraphics[height=10cm]{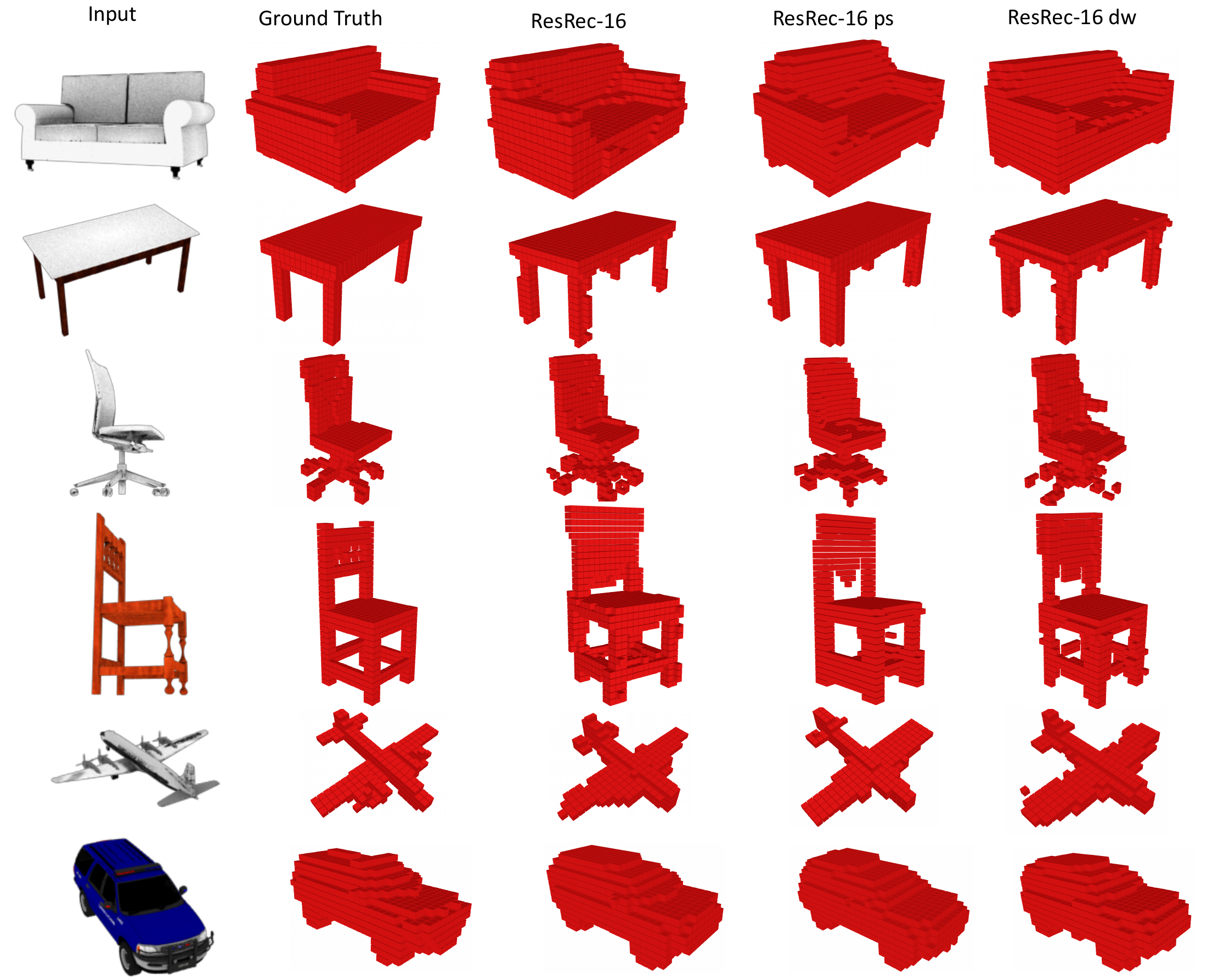}
\caption{Qualitative results of reconstruction using ResRec with standard, pseudo and depthwise 3D convolutions.}
\label{fig:rec-viz}
\end{figure}

We also show some qualitative results of using all $3$ types of 3D convolutions in figure~\ref{fig:rec-viz}. Interestingly, although their quantitative performances are almost the same, different types of convolutions yields different styles of reconstructions. Pseudo-3D delivers smooth surfaces but tends to ignore some details. 3D depthwise generally produces rough surfaces however keeps the details better. Standard 3D convolution seems to be a trade-off solution for balancing smoothness and details.

\subsubsection{Depth matters in decoders}


\textbf{}

\begin{table}[h!]
\centering
\caption{Performances of 3D-R2N2s using different decoders. We also copy results for Rec-16 and ResRec-16 in last two lines for convenience of comparison.}
\begin{tabular}{c|c|cc}
\hline
encoder & decoder  & mIoU \\ 
\hline
3D-R2N2 encoder & 3D-R2N2 decoder & $\ \ 55.1\ \ $ \\
3D-R2N2 encoder & ResRec-16 decoder & $\ \ 59.5 \ \ $ \\
\hline
Rec-16 encoder & Rec-16 decoder & $\ \ 63.4\ \ $ \\
ResRec-16 encoder & ResRec-16 decoder & $\ \ 63.1\ \ $ \\
\hline
\end{tabular}
\label{table:depth-matter-in-decoder}
\end{table}

We notice that 3D-R2N2~\cite{choy20163d}, which was the state-of-the-art at that time, has a lower mIoU comparing to all other seemingly concise models. We show in experiments that model's depth has a major influence on decoder's performance. Both using residual blocks, the ResRec-16 decoder has more layers than 3D-R2N2 decoder. Keeping the encoder unchanged, by switching to the ResRec-16 decoder, the modified 3D-R2N2's mIoU raises more than $4\%$ as suggested in table~\ref{table:depth-matter-in-decoder}.

Though we've seen significant improvement with the ResRec-16 and Rec-16 decoder in table~\ref{table:depth-matter-in-decoder}, the modified 3D-R2N2s' performances are still almost $ 4\%$ behind ResRec-16 and Rec-16, leading to the conclusion that ResNet50 is also a better encoder than 3D-R2N2's encoder. Notice that the 3D-R2N2 encoder does intergrate residual block into it. But it's shallow comparing to ResNet50 and it's not pretrained on ImageNet.

This result confirms our intuition that, in current stage, we do need deeper 3D ConvNets to conduct 3D vision tasks. And 3D depthwise convolution can help us build deeper 3D ConvNets under when computing resources are constraining the model size.

\section{Conclusion}\label{conclusion}
In this paper, we have proposed the use of 3D depthwise convolutions which improves 3D ConvNets' efficiency by reducing the number of parameters. We've shown that 3D depthwise convolution operation has comparable performance to standard 3D convolution operations in classification task with $\sim 95\%$ fewer parameters. We've also demonstrated the potential of using it to improve 3D neural nets' performance in other 3D vision tasks like 3D Reconstruction and got decent results with minimum number of parameters in convolutional layers. And the experiment has further indicated that deeper 3D ConvNets are indeed needed in 3D vision tasks, where 3D depthwise convolution can help.

\clearpage

\bibliographystyle{splncs}
\bibliography{egbib}
\end{document}